%% file: neurips_2020.tex
\documentclass{article}

\usepackage[final, nonatbib]{neurips_2020}

\usepackage[utf8]{inputenc} % allow utf-8 input
\usepackage[T1]{fontenc}    % use 8-bit T1 fonts
\usepackage{hyperref}       % hyperlinks
\usepackage{url}            % simple URL typesetting
\usepackage{booktabs}       % professional-quality tables
\usepackage{amsfonts}       % blackboard math symbols
\usepackage{nicefrac}       % compact symbols for 1/2, etc.
\usepackage{microtype}      % microtypography
\usepackage{wrapfig,lipsum}
\usepackage{enumitem}
\usepackage{pifont}
\usepackage{textcomp}
\usepackage{subfigure}
\usepackage{amsmath,amssymb}

\usepackage{varwidth}
\usepackage{booktabs}
\usepackage{arydshln}
\usepackage{graphicx}
\usepackage{multirow}
\usepackage[T1]{fontenc}
\usepackage{pifont}

\usepackage{bm}
\usepackage[dvipsnames,table,xcdraw]{xcolor}
\usepackage[font=footnotesize]{caption}

\newcommand\nl[1]{\emph{``#1''}}

\newif\ifcomments
\commentstrue 
\ifcomments
    \providecommand\at[1]{[\textcolor{blue}{{AT: #1}}]}
    \providecommand\jb[1]{[\textcolor{red}{{JB: #1}}]}
    \providecommand\ot[1]{[\textcolor{violet}{{OT: #1}}]}
    \providecommand\pc[1]{[\textcolor{violet}{{PC: #1}}]}
    \providecommand\yg[1]{[\textcolor{purple}{{YG: #1}}]}
\else
    \providecommand\at[1]{}
    \providecommand\jb[1]{}
    \providecommand\ot[1]{}
    \providecommand\pc[1]{}
    \providecommand\yg[1]{}
\fi
\newcommand\comment[1]{}
\newcommand\bfemph[1]{\textbf{\emph{#1}}}

\newcommand\ruletaker{\textsc{RuleTaker}}
\newcommand\twentyquestions{\textsc{20Q}}
\newcommand\counting{\textsc{Counting}}
\newcommand\hypernyms{\textsc{Hypernyms}}
\newcommand\roberta{\textsc{RoBERTa}}
\newcommand\robertal{\textsc{RoBERTa-Large}}

\newcommand\zeroshot{\textsc{RuleT.+20Q}}
\newcommand\esim{\textsc{ESIM}}
\newcommand\statementOnlyLangSelectivity{\textsc{Statement-only-language-selectivity}}
\newcommand\statementOnlyNoContext{\textsc{Statement-only-no-context}}
\newcommand\statementOnly{\textsc{Hypothesis-only}}
\newcommand\softReasoningOnly{\textsc{Explicit Reasoning}}
\newcommand\implicitHypernyms{\textsc{Implicit Reasoning}}
\newcommand\controlledDS{\textsc{Imaginary}}

\newcommand\countingstatementonly{\textsc{Hypothesis-only}}
\newcommand\countingexperiment{\textsc{Counting}}

\newcommand\truehypernym{\texttt{relevant hypernym}}
\newcommand\falsehypernym{\texttt{distractor hypernym}}
\newcommand\trueproperty{\texttt{relevant property}}
\newcommand\falseproperty{\texttt{distractor property}}
\newcommand\subjectdistractor{\texttt{distractor subject}}
\newcommand\predicatedistractor{\texttt{distractor predicate}}

\newcommand\context{explicit knowledge}

\newcommand\memberfact{member fact}

\newcommand\quantityfact{quantity fact}
\newcommand\Quantityfact{Quantity fact}

\newcommand\manualset{\textsc{Multi-skill-set}}

\usepackage{quoting}

\newcommand{\eat}[1]{}

\title{Leap-Of-Thought: Teaching Pre-Trained Models to Systematically Reason Over Implicit Knowledge }

\newcommand{\emailsize}{\fontsize{12pt}\times}
\author{Alon Talmor$^{1,2}$ ~~ Oyvind Tafjord$^{1}$ ~~
Peter Clark$^{1}$ ~~
Yoav Goldberg$^{1,3}$ ~~
Jonathan Berant$^{1,2}$ \\
\mbox{}\\
$^1$The Allen Institute for AI \\
$^2$Tel-Aviv University, ~
$^3$Bar-Ilan University \\
\emailsize{\texttt{\{alont,oyvindt,peterc,yoavg,jonathan\}@allenai.org}}}

\date{}

\begin{document}
\maketitle
\input{00_abstract}
\input{01_intro}

\input{02_motivation}
\input{03_method}

\input{04_experiments}

\input{04a_taxonomy}

\input{04b_analyzing_systematicity}

%\input{04ab_synthetic}
\input{04c_counting}

\input{04d_combining_skills}

\input{05_related_work}

\input{06_discussion}

\input{07_statement_of_broader_impact}
\input{08_acks}
\bibliography{all}
\bibliographystyle{unsrt}
\input{10_supplementary}

\end{document}

%% file: 00_abstract.tex
\begin{abstract}

To what extent can a neural network systematically reason over symbolic facts? Evidence suggests that large pre-trained language models (LMs) acquire some reasoning capacity, but this ability is difficult to control. Recently, it has been shown that Transformer-based models succeed in consistent reasoning over explicit symbolic facts, under a ``closed-world" assumption. However, in an open-domain setup, it is desirable to tap into the vast reservoir of implicit knowledge already encoded in the parameters of pre-trained LMs. In this work, 
%\pc{If you wanted to be drastic and concise, you could simply start the abstract here and cut the earlier bit! :)} 
we provide a first demonstration that LMs can be trained to reliably perform systematic reasoning combining {\it both} implicit, pre-trained knowledge and explicit natural language statements.  To do this,
we describe a procedure for automatically generating datasets that teach a model new reasoning skills, and demonstrate that models learn to effectively perform inference which involves implicit taxonomic and world knowledge, chaining and counting. 
Finally, we show that ``teaching'' the models to reason generalizes
beyond the training distribution: they successfully compose
the usage of multiple reasoning skills in single examples. Our work paves a path towards open-domain systems that constantly
improve by interacting with users who can instantly correct a model by adding simple natural language statements.

\end{abstract}

%% file: 01_intro.tex
\section{Introduction}

A longstanding goal of artificial intelligence is to develop systems that continuously accumulate knowledge by consuming facts and rules about the world and reasoning over them \cite{mitchell2018never,tunstall2010true,Davis1977InteractiveTO}. Recently \cite{clark2020transformers}, it has been shown that Transformers \cite{vaswani2017attention} are an effective architecture for this goal, as they can be trained to reason over knowledge expressed as natural language statements.

However, modern neural networks for language do not store
%\yg{why do we care how it is stored? we care about how to interact with it (i think this comments unlike the others is not a in-place small change, and may require larger change, if you decide its worthwhile.)}
knowledge symbolically. Instead, substantial amounts of knowledge are encoded in their parameters by pre-training on large corpora with a language modeling (LM) objective \cite{peters2018elmo,devlin2019bert,liu2019roberta, radford2018improving}.
Moreover, although LM-based models do exhibit certain reasoning abilities \cite{richardson2019does,talmor2019olmpics}, these abilities are not systematic and are difficult to control. 
Thus, for real-world open-domain applications, it is imperative that models reason consistently over both \emph{explicit} input statements and \emph{implicit} knowledge that is already encoded by the network. 

\begin{wrapfigure}[17]{R}{0.6\columnwidth}
  \includegraphics[width=0.6\columnwidth]{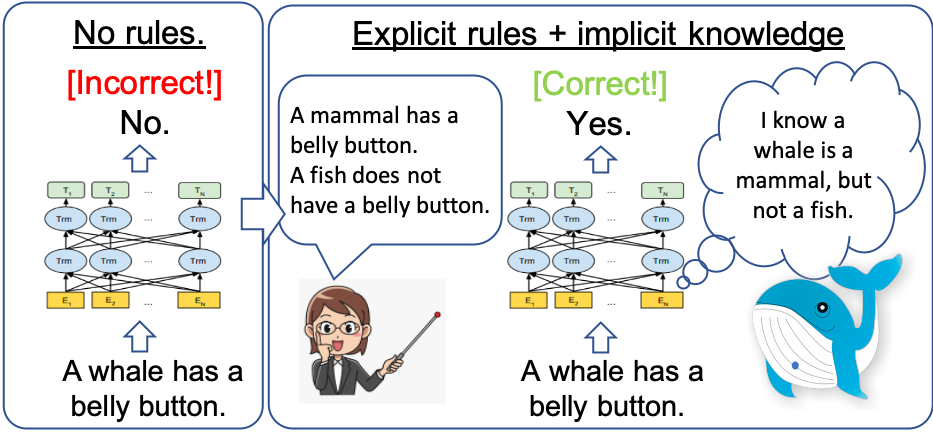}
  \caption{
  %JB: for brevity
  %Combining explicit and implicit knowledge: 
  The model is wrong when asked whether \nl{A whale has a belly button}. 
  %However, given the rules \nl{A mammal has a belly button} and \nl{A Fish does not have a belly button}, %the model uses implicit knowledge to answer correctly.
  However, if a user tells the model the explicit rule \nl{A mammal has a belly button}, the model combines this on-the-fly with its implicit knowledge that \nl{A whale is a mammal}, and arrives at the right conclusion (without re-training).}
  %the model instantly, from a single example, reverses it's prediction and "changes it's mind". }
  ~\label{fig:intro-fig}
\end{wrapfigure}

%\at{Perhaps a more practical reason to combine implicit knowledge with explicit is not because of the way transformers are trained, but rather the fact that in real life application we rarely have all rules required for reasoning in the exact same language as the question at hand}

In this work, we develop models that reason over implicit knowledge and explicit natural language statements in a systematic manner. 
%We envision a model developer that would like to endow a model with the ability to perform a certain type of inference, where \emph{knowledge} is (partially) encoded by the model internally, but \emph{rules} are given explicitly as text.
%\pc{The knowledge/rules terminology makes me whince a bit... rules are knowledge, yes?. You could simply delete this 2nd sentence, the example carries the point.}
Consider the example in Figure~\ref{fig:intro-fig}. The model needs to determine whether \nl{A whale has a belly button}, but answers incorrectly since this particular knowledge nugget is unknown to the model. However, if a user tells the model explicitly that \nl{A mammal has a belly button}, the model can combine this statement with the its implicit knowledge that \nl{A whale is a  mammal}, and make the right inference on the spot. Thus, we want the model to systematically handle such cases that combine implicit knowledge with input natural language statements. 
%\pc{Perhaps distinguish from RTE/NLI, which already does this?}

%\jb{if we have a counting example, we'll add the commented out sentence}
%Figure~\ref{fig:bla} (right) shows another type of inference that involves counting, where the model must learn whether a certain fact is possible by counting over facts that are either given explicitly, or known a-priori by the model.

We train our models by automatically generating examples that illustrate the expected types of inference. Because the knowledge of pre-trained models comes from real world text, we test this knowledge by generating examples using true facts and rules from multiple information sources. 
%Training examples include cases where all the information is provided explicitly as natural language statements, where the model can focus on learning the inference rule, and cases where implicit knowledge must be combined for the correct inference to take place.

We focus on two types of high-level reasoning: (a) inference that combines implicit taxonomic knowledge (hypernymy, meronymy, etc.) with explicit natural language rules, and (b) inference that requires counting over both implicit and explicit facts, and checking whether a certain count was reached. In both cases, we observe that models can be trained to reason over implicit and explicit knowledge. Importantly, model performance can be explained by its prior knowledge: inference is successful when the necessary knowledge exists in the model, and fails when it is missing. 

Last, we show that training the models to perform inference generalizes
\emph{beyond} their training distribution. Specifically, we endow a pre-trained
LM with multiple inference capabilities independently, and show that it can
handle examples that require composing multiple inference types, even when these
%combinations 
do not appear at training time. Thus, one can gradually
improve the inference capabilities of a model and expect generalization.

%%% THIS IS THE EXAMPLE FROM SECTION 2, MOVED HERE TO PUSH IT CLOSER TO 

Our work paves a path towards systems that constantly improve by interacting with users: when a user spots an error, they can fix that error by providing a single statement in natural language that will allow the model to apply its acquired inference skills and reach the right conclusion.
If the system can successfully retrieve this statement in future interactions, it will fix not only the current mistake, but also future ones. This can be viewed as a form of ``one-shot learning'' that improves the model on-the-fly without further training, unlike most current work that relies on data collection and re-training for fixing model errors.
%This approach provides a natural and seamless process, compared to current practice of adding training examples and re-training each time a new rule is required. 
All our code and data is publicly available at \url{http://github.com/alontalmor/LeapOfThought}.

\comment{
\begin{itemize}

%background

%Problem
\item However, when something is wrong, the main way to fix things is to add/change data and re-train. This is perhaps feasible for developers (alebeit slow), but not possible for a user. Consider an example (whales examples)... We would like to let users talk to the model in natural langauge and provide a rule that will systematically correct this and many other related errors.

\item \at{One possible approach is to hand the model a set of explicit facts/rules in the context and have it logically reason on them to arrive at the current answer (cite ROVER). However, this requires KB+IR rules for every possible piece of knowledge to be available and correct, which is infeasible. }

% In this work

% Empirical investigation
We teach our models to perform two types of inference. First, to combine implicit taxonomic knowledge that is common in LMs with quantified rules... describe what we do here. Second, we train our model to perform numerical reasoning by performing an approximation of counting over implicit and explicit facts. Finally, we show taht....

%Ramification
Our work shows that...

\end{itemize}
}

%% file: 02_motivation.tex
%\section{Combining Implicit and Explicit Knowledge: A Motivating Example}

\section{A Motivating Example}
\label{sec:motivation}

\begin{wrapfigure}[19]{r}{0.38\columnwidth}
\vspace{-40pt}
\includegraphics[width=0.38\columnwidth]{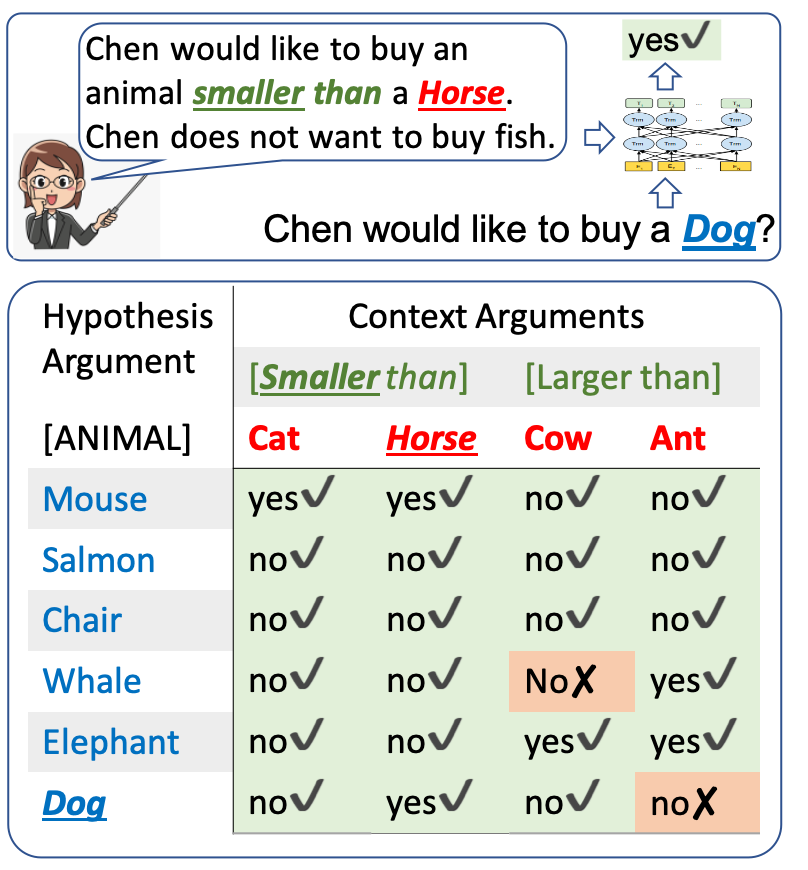}
\caption{Our motivating task. The top box shows a single example. In the bottom we systematically replace arguments matching the color of the bold underlined words in the top example. The text in the table corresponds to model predictions, and the color and $\checkmark$ indicate a correct prediction. 
%JB: delete for brevity?
%Solving this task requires combining implicit knowledge of hypernymy and size comparison with the explicit statements in the context.\at{shorten?}
}
\vspace{40pt}
~\label{fig:motivational-example}
\end{wrapfigure}

%\pc{I was expecting this section to say "let's try a simple way of learning implicit + explicit, but we find it doesn't work (hence the rest).", but it reads as "we did a quick test and it works!" So can't we just stop and go home? :) Need to rephrase as we discussed, e.g., "we find a surprising result.... can we now capitalize on this in a systematic way?" etc.}
%\yg{the word "simple"/"simply" is used way to much in this text.}
%\yg{Maybe open with: 
We begin by demonstrating that combining reasoning over explicit input with reasoning over implicit LM knowledge can ``emerge'' by training a Transformer on a dataset that requires each skill individually. 
%We begin our research by exploring the potential for combining implicit and explicit knowledge while performing a simple experiment that checks whether reasoning can ``emerge" simply by training on existing datasets. 
We fine-tune \roberta{} \cite{liu2019roberta}, on binary (yes/no) question answering tasks from two datasets (using standard multi-task training): (a) 50K examples from \textsc{Twenty Questions} (\twentyquestions{}),\footnote{\url{https://github.com/allenai/twentyquestions}} a question answering (QA) dataset which includes questions such as \nl{Does an aircraft fly?} (\texttt{true}) and \nl{Do everyone have an alarm?} (\texttt{false}). This teaches the model to retrieve real world facts from its internal implicit knowledge; and
%which teach the model to retrieve (and combine?) facts from its internal implicit knowledge} which strengthen the model's commonsense and world knowledge; 
(b) 100K examples from the \textsc{RuleTaker} \cite{clark2020transformers} reasoning dataset, teaching the model to reason over a set of assertions explicitly provided as natural language statements. 
%We fine-tune \roberta{} in a multi-task fashion, by combining examples from both dataset in a random order, using a lower that usual learning rate (5e-6).

We evaluate this model on a task that requires combining implicit knowledge about the sizes of animals (known to exist in \roberta{} \cite{talmor2019olmpics}) and an animal taxonomy, with explicit reasoning over natural language statements.
Figure~\ref{fig:motivational-example} illustrates the setup. The model needs to determine if a hypothesis of the form \nl{Chen would like to buy a [ANIMAL]} is true, where the slot \emph{ANIMAL} is replaced with some object. The model also observes \context{} that specifies the desired size of the animal and that it must not be a fish. Over 24 animal pairs, the model obtains 91.6\% accuracy, successfully combining implicit knowledge with the given statements in ways that were not observed during fine-tuning. Variations of this task work equally well.

%While evaluated the model on various manually crafted \nl{toy} examples, strikingly, we discovered that the model is able to successfully combine multiple reasoning abilities, not observed in the fine-tuning sets, in the same example.
%Figure~\ref{fig:motivational-example} shows one such example. The hypothesis contains the statement for which the model should predict yes or no. The context contains a set of two rules. Both hypothesis and rules contain arguments, that are replaced with values displayed in the table below. Model predictions are shown in table-cells matching the hypothesis and context arguments.
%Overall the model achieves 91.6\% on this out of distribution set of examples, correctly asserting that a chair (as well as other objects inserted) is not an animal, a Salmon (and other fish types tested) is a fish. Correctly reasoning about the various sizes, as well as performing a conjunction between both rules. 

%\yg{We show that while each training example contained either a set of rules \emph{or} implicit knowledge, the model learned to effectively combine both.}\yg{(this is said below, but I tried to make things more direct.)}
%We now evaluate whether the model is able to reason over knowledge that is represented in the network parameters and natural language statements. \jb{explain how we test this and mention two numbers that show that this thing is doing better than expected} \jb{refer to cool examples}

This experiment shows that although training examples did not require reasoning over explicit statements \emph{and} implicit knowledge, the model learned to effectively do so. While exciting, the procedure is still an ``alchemy''. Can we make it more systematic and more controlled, and get a better handle on its capabilities and limitations? The rest of the paper explores these questions.

%Surprisingly, this experiment suggests that models can be trained to perform various reasoning skills independently, and then successfully combine those skills. But how can we make this reasoning systematic? And can we devise a procedure for injecting new skills into the model? We turn to this next.

%Our analysis highlights \at{suggests? implies? we didn't really show it yet} that injecting knowledge and reasoning abilities into models \jb{ref ankit and mor} on independent benchmarks can naturally lead to models that compose their abilities and generalize in a surprising manner. But does the model reason in a systematic manner? Can we improve systematicity by explicitly training the model to apply particular inference rules? This is the focus of the next sections.

%% file: 03_method.tex
\section{Method}
\label{sec:method}

We describe a general procedure for creating models that can perform inference over explicit and implicit knowledge. The process includes automatic data
% information
generation from existing sources, followed by standard supervised training. \S\ref{subsec:taxonomy} and \S\ref{subsec:counting} show
% describe
two instantiations of this procedure.
%We then describe two instantiations of this procedure in \S\ref{subsec:taxonomy} and \S\ref{subsec:counting}.

%Consider a developer\yg{this is a somewhat abrupt jump to focus on "developers" which i am not sure is needed / beneficial.} interested in a model that can apply certain inference rules over its prior knowledge. We describe a procedure for accomplishing this through automatic data generation \jb{ref ankit and mor}, and demonstrate its implementation in subsequent sections. 

% \yg{proposal: change "hypothesis" to "assertion" throughout?}
\paragraph{Definitions}
Our goal is to endow a model with the ability to perform certain \emph{inference types}. 
%\pc{It'd be nice to crisply define "inference type"....although not easy.... Perhaps instead of the below sentence, say: "
An {\it inference type} describes how two or more natural language statements combine logically. For example, we define the
\textsc{hypernymy} inference type 
%as the a syllogism 
with the assertion  %using the pattern \texttt{`if A is a type of B, and B has property P, then A has property P'}"}
%For example, syllogism is an inference type that can be described as 
\texttt{`if A is a type of B, and B has property P, then A has property P'}. Similarly, we will later use \textsc{approximate counting}, and \textsc{meronymy} as additional inference types. 
To this end, we automatically generate training examples, where each example includes (a)  a \emph{hypothesis}, i.e., a textual statement that is either true or false (\nl{A whale has a belly button}), and (b)
%a 
\emph{\context{}}, which
is a list of textual statements. Statements can either be \emph{facts}, that is, describe a property of a particular entity (\nl{John Lennon is a member of The Beatles}), or \emph{rules}, that is, describe a property of some class (\nl{Mammals have belly buttons}).
%Context s are constructed 
The \context{} is constructed such that the truth value of the hypothesis cannot be inferred from the  \context{} alone, but also requires some knowledge to be encoded by the LM a-priori.
For example, the \context{} might not include the rule \nl{A whale is a mammal}, necessary for deducing that they have belly buttons (Figure~\ref{fig:intro-fig}).

\paragraph{Data generation}
% data sources and data type
Our primary motivation is to develop models that work in an open domain environment with real world facts. Thus, 
we automatically generate data by sampling from existing knowledge sources: \textsc{ConceptNet} \cite{speer2017conceptnet}, \textsc{WordNet} \cite{fellbaum1998wordnet} and \textsc{Wikidata} \cite{vrandecic2014wikidata}.
We sample (\texttt{subject}, \texttt{predicate}, \texttt{object}) triples  that are known to be either \texttt{true} or \texttt{false}.
Pseudo-language statements are then generated using manually constructed templates for each \texttt{predicate}.\footnote{The full list of predicates used is available in the supplementary material.}
%For hypotheses and facts, we generate pseudo-language statements by manually constructing a fixed template for each \texttt{predicate}.
%JB: this was already defined above.
% example format / structure description
%Each example includes a statement text which can be true or false, and a rules context which consist of a set of rule sentences.
% describe the pseudo-language generation: 
%For each \texttt{predicate} type we manually generate pseudo-natural language phrases by using predefined templates. 
For example, for true statements such as (\texttt{chicken}, \texttt{has}, \texttt{feathers}) we generate \nl{A chicken has feathers.}; for false statements, we generate \nl{A chicken does not have horns}.
%Certain predicates (corresponding to taxonomic relations such as hypernymy and meronymy) can be mapped to rules: the true triple (\texttt{bird}, \texttt{has}, \texttt{feathers}) is mapped to the rule \nl{If something is a bird, then it has feathers}, and similarly the false triple (\texttt{bird}, \texttt{has}, \texttt{horns}) is mapped to  \nl{If something is a bird, then it does not have horns}.
In all examples, the order of the statements in the  \context{} is random and the number of true and false hypotheses is equal.

\paragraph{Training}
Once examples are generated, we fine-tune a pre-trained LM, specifically \robertal{}  \cite{liu2019roberta}, on this dataset.
The inputs and outputs are modeled in the standard manner \cite{devlin2019bert}: The input is given as a list of tokens  \texttt{`[CLS]  \context{} [SEP] hypothesis [SEP]'}, and the contextualized representation of the \texttt{[CLS]} token is linearly projected down to two logits, and passed through a softmax layer to obtain the probabilities that the hypothesis is true or false. We train by minimizing the binary cross-entropy loss, and evaluate models using accuracy.
%Because the labels of generated datasets are balanced, a random baseline obtains 50\% accuracy. 
%\ot{Note: I updated description to involve two logits since that's what we're doing now, rather than a sigmoid, although it shouldn't matter}
%We train \roberta{} to predict true/false (i.e., binary classification) for each question statement. Questions are supplied to \roberta{} as: [CLS] context [SEP] statement [SEP], where context is the theory (facts+rules, expressed in psuedo-language) and statement is the fact for which correctness is to be determined. The [CLS] output token is projected to a single logit. A logit score of >0 is treated as predicting true, otherwise the answer is false.
%Training is performed using cross-entropy loss. For evaluation, we measure accuracy. \at{(The test data has an equally balance of TRUE/FALSE answers, hence the baseline of random guessing is 50\%). ? }

In addition, To investigate the importance of pre-trained contextualized representations, we use ESIM \cite{chen2017enhanced} over non-contextualized \textsc{GloVe} \cite{pennington2014glove} representations as a baseline architecture, as it is known to provide a strong model when the input is a pair of text fragments \cite{chen2017enhanced}. We adapt the architecture to the mutli-choice setup using the procedure proposed by \cite{zellers2018swag}, with two choices for `yes' and `no'. 
Last, motivated by the results in \S\ref{sec:motivation}, we evaluate whether the \zeroshot{} model, described in \S\ref{sec:motivation}, can utilize implicit knowledge even without being directly trained for this goal. 

\comment{
\begin{itemize}

\jb{would be good to define stuff here. What I have in my head is that there is something like the `type of inference' that we want a NN to be able to perform. This is something like `if A is a hyponym of B and B has property C then A has property C'. For each such inference type we will create a training set. The goal of the training set is to teach the model the rule, such that it can be applied easily. So there is the `type of inference' and then its application. I think it would be good to talk about having (a) questions (b) facts (c) rules. I know the counting thing is not exactly a rule, but I think it is close enough.}

\jb{should we talk here about the fact that we want to teach the model both how to do the inference and also to do this over implicit knowledge, and that's why we combine examples that are both explicit and implicit, so that it is easier? We can perhaps show that if you train only with implicit it does not work}

\jb{talk about the models and training very shortly}

\jb{Is there anything about the data generaion that should be here or will it be in the experiments?}

\item Our setup is Question Answering (QA) with Boolean answer type. 

\item One possible impact: Using this method we can correct a model that errors K times on some eval-set with M << K rules. The model predicts "NO" for "Cats are maid of matter". If we are able to teach it "All animals are made of Matter", we can correct \textsc{number of animals} examples with only 1 rule. 

\end{itemize}
}

%% file: 04_experiments.tex
\section{Experiments}
\label{sec:experiments}
We instantiate our approach over two inference types. First, correcly applying implicit taxonomic knowledge (\S\ref{subsec:taxonomy}). Second, %approximate
counting over explicit and implicit facts (\S\ref{subsec:counting}). Additionally, we analyze our results and show that model success can be reliably predicted by probing the background knowledge encoded in the pre-trained LM (\S\ref{subsec:systematicity}).

%% file: 04a_taxonomy.tex
%\subsection{Can a Language Model combine taxonomic knowledge with soft-reasoning?}
%\subsection{Leveraging implicit knowledge of hypernym relations}

\subsection{Implicit Knowledge of Taxonomic Relations}
\label{subsec:taxonomy}
Pre-trained LMs have been shown to capture substantial amounts of 
taxonomic knowledge such as hypernymy (\texttt{A is a type of B}) and meronymy (\texttt{A is part of B}) \cite{richardson2019does}.
To test whether this knowledge can be leveraged for reasoning, we create examples that can be solved only by integrating this implicit knowledge. %For example, Given a hypothesis \nl{A \textbf{cat} is purple} and a  \context{} \nl{All \textbf{animals} are purple}, the model must implicitly use the fact that \nl{cat} is a hyponym of \nl{animal} in to predict the right answer.

\paragraph{Data Construction}

\begin{wrapfigure}[13]{R}{0.4\columnwidth}
\vspace{-12pt}
\includegraphics[width=0.4\columnwidth]{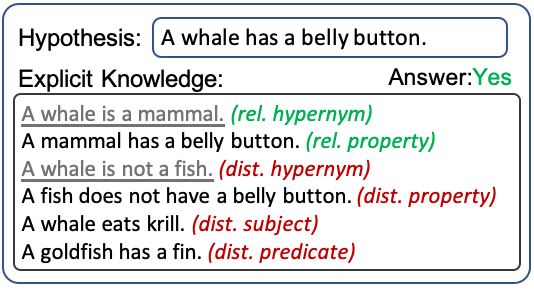}
\caption{Outline of a taxonomy example. The purpose of each relevant and distractor rule is in parenthesis. 
%The relevant hypernym and property rules are in bold. 
%in this case the positive pair and the negative pair servers as distractor. 
Underlined hypernym rules are removed in \implicitHypernyms{}. }
\vspace{12pt}
~\label{fig:hypernyms-example}
\end{wrapfigure}

Figure~\ref{fig:hypernyms-example} illustrates the main components of an example. We describe the generation process, based on triples from \textsc{ConceptNet} and \textsc{WordNet}.

To generate a positive example (\texttt{true} label), we sample a \truehypernym{} rule, e.g., ((\texttt{whale}, \texttt{is a}, \texttt{mammal}), \texttt{true}). We then find a \trueproperty{} of the hypernym \texttt{object}, e.g., ((\texttt{mammal}, \texttt{has a}, \texttt{belly button}), \texttt{true}). 
We apply the hypernymy inference type (downward monotonicity): %\pc{"downward monotonicity" $\rightarrow$ "the hypernymy inference type" or "the hypernymy syllogism"} 
(\texttt{if A is a B and B has property C, then A has property C}) to deduce the logical conclusion which will become the hypothesis, e.g., ((\texttt{whale}, \texttt{has a},
\texttt{belly button}), \texttt{true}). 

To add distractors, we create similar but irrelevant statements by: (a) randomly replacing the \texttt{subject} of the \trueproperty{} to create a \falseproperty{}, e.g., ((\texttt{fish}, \texttt{has a}, \texttt{belly button}), \texttt{false}). 
(b) Create a \falsehypernym{} by combining the subject of the hypothesis and the subject of the \trueproperty{}: e.g., ((\texttt{whale}, \texttt{is a}, \texttt{fish}), \texttt{false}), 
(c) add a \subjectdistractor{} by randomly sampling a different rule about the hypothesis \texttt{subject}, and 
(d) add a \predicatedistractor{} by randomly sampling a different rule with the hypothesis predicate. Thus, the  \context{} contains the relevant hypernym and property, and four distractors.

Negative examples are created from positive examples by using the \falseproperty{} (\nl{a fish does not have a belly button}). We sample a hypernym rule, such that the object matches the subject of the \falseproperty{} e.g. ((\texttt{salmon}, \texttt{is a}, \texttt{fish}), \texttt{true}). We then apply the hypernymy inference type to obtain the false hypothesis ((\texttt{salmon}, \texttt{has a}, \texttt{belly button}), \texttt{false}). In the negative example the roles of the relevant hypernym and property, and distractor hypernym and property are reversed. The distractor subject and predicate are sampled independently. 

%and again apply downword monotone to produce the a false statement:   (\texttt{\nl{Salmon}}, \texttt{\nl{has a}}, \texttt{\nl{Belly button}}, false). Both positive and negative examples contain two pairs of positive and negative hypernyms and properties rules, one serves as relevant pair, and the other as distractor.
 
%In order to correctly determine if the statement is true or false, the model must choose the relevant hypernym and property pair.
We generate 30,906 training examples using this procedure. We create  development and test sets, 1,289 examples each, where the subjects and objects 
%in the training set  
are \emph{disjoint} from the ones in the training set.
%development and test sets. 
In 50\% of the training examples, we remove the relevant and distractor hypernyms to teach the model to use implicit knowledge. We find that without this step, the model always predicts false if a necessary hypernym is missing, performing well only on the \softReasoningOnly{} setup. In 20\% of the training examples we remove all distractors.\footnote{The training set also includes 2,864 examples that contain only a hypothesis, half of which are true hypernym rules (\nl{a whale is a mammal}) and half are false hypernym rules (\nl{a whale is a fish}). This is useful for analyzing what prior knowledge a LM has, as we show in \S\ref{subsec:systematicity}.} In the test set, we remove some of the statements in the  \context{}, based on the experimental setting, as explained below. For example, we remove all hypernyms to test the use of implicit knowledge.

%To test the model for implicit pre-trained knowledge, we remove hypernym rules in some setups, forcing the model to determine what is the relevant property based on its prior implicit knowledge.

%We add a randomly chosen \subjectdistractor{} rule matching the subject of the statement and a \predicatedistractor{} matching the statement predicate as distractors.

%For training, we remove the \truehypernym{} and \falsehypernym{} in 50\% of the examples, and remove all irrelevant rules in 20\% of the examples \jb{all distractors?}. 5\% of the \truehypernym{}s and 5\% of the \falsehypernym{}s are added as statement in examples with no rules \jb{no  \context{}?}, to adapt the model for testing its knowledge of these hypernym relations.  \jb{the 5\% thing is unclear (I did not understand)}

%We split examples such that objects in training and evaluation are disjoint. We generate total of 31,500 training examples, and 1,888 evaluation examples. 
\noindent
\bfemph{Meronyms (zero-shot)}
%\paragraph{Meronyms (zero-shot)}
To test whether the inference type learned by the model generalizes to other inference types, we used the same procedure to create 2,528 test examples using a \textsc{meronymy} inference type (meronym transitivity),
i.e. \texttt{`if A has part B and B has part C, then A has part C'} or \texttt{`if a hand has a cell and a cell has an atom, then a hand has an atom.'}. 
There is no training set in this setup.

\paragraph{Experiments}
We evaluate our model in three different setups:
\begin{itemize}[leftmargin=*,topsep=0pt,itemsep=2pt,parsep=0pt]
\item[] \statementOnly: The model is given only the hypothesis without the  \context{}. This tests whether the model already knows the answer without the need for inference.
\item[] \softReasoningOnly: The model is given the hypothesis and  \context{} with hypernym rules, thus the model needs to perform inference over explicit natural language statements.
\item[] \implicitHypernyms: The model is given the hypothesis and  \context{} \emph{without} hypernym rules, and is forced to choose the correct answer based on its implicit knowledge.
\end{itemize}

%\yg{If space, i'd add something like ``We expect to see low scores under conditions X, highest score under condition and Y, and good scores for condition Z. The gap between Y and Z is the ability of the model to perform..''}

\paragraph{Models}
We compare the following models: (a) \roberta{} fine-tuned on the generated dataset, (b) The \zeroshot{} model (see 
\S\ref{sec:method}), and (c) \esim{} trained on the generated dataset without pre-training (testing the value of pre-trained contextualized representations).

\paragraph{Results}
We show the results in Table~\ref{tab:hypernym-meronym}.
%\at{talk about \zeroshot{} \esim{}  when i get the results...}
On \statementOnly{} evaluation examples, \roberta{} achieves 65.2\% accuracy, substantially higher than random (50\%), suggesting it knows some of the answers even without the \context{} (if we train only on \statementOnly{} examples, accuracy on  \statementOnly{} test examples is slightly higher at 69.7\%). \softReasoningOnly{} shows that when the relevant hypernym and property are both present in the  \context{}, \roberta{} achieves near perfect performance, easily solving 2-hop reasoning.  Finally, the accuracy of  \implicitHypernyms{} is 88.8\%, substantially higher than  \statementOnly{}.  This suggests the model is able to correctly apply its implicit knowledge of hypernyms to select the relevant property rule, effectively performing 2-hop reasoning with one hop done internally.

\begin{wraptable}[12]{R}{0.55\columnwidth}
\vspace{-13pt}
\begin{center}
\resizebox{0.55\columnwidth}{!}{
\begin{tabular}{l|c|c|c|c}
Model  $\rightarrow$ & \multicolumn{3}{c|}{\roberta{}}    & \esim{}  \\
\toprule
Train-set $\rightarrow$  & Hyper. & Hyper.  & \zeroshot{} & Hyper.  \\ 
\midrule
Test-set $\rightarrow$ & Hyper. & Mero.   & Hyper. & Hyper.   \\
\midrule
\statementOnly      &   65.2  &  70.8   &  65.4   & 61.3      \\
\softReasoningOnly  &   99.7  &  99.4   &  98.4   & 79.8      \\
\implicitHypernyms  &   \textbf{88.8}  &  \textbf{86.9}   &  \textbf{79.1}   & \textbf{76.1}      \\
\bottomrule
\end{tabular}
}
\end{center}
\caption{Test set results for reasoning over hypernymy and meronymy relations. The models learn to reason with implicit rules, significantly improving on the hypothesis-only baseline, some in zero-shot.}
\vspace{13pt}
\label{tab:hypernym-meronym}
\end{wraptable}

We evaluate performance of \zeroshot{} to determine if combining implicit and explicit knowledge emerges even without direct training.\footnote{ Clark et al. \cite{clark2020transformers} tested whether \roberta{} can perform reasoning in the few-shot setup, without further fine-tuning on additional auxiliary datasets, and found that models perform poorly (figure 4 in their paper).}
The model, although trained on a different distribution for explicit reasoning (the \textsc{RuleTaker} dataset), achieves 98.4\% on \softReasoningOnly{}. Surprisingly,
without being trained to perform implicit reasoning, accuracy improves from $65.4 \rightarrow 79.1$ when given the  \context{} without hypernym rules (but still lower than $88.8$).
\esim{}, trained with \textsc{GloVe} pre-trained word embeddings, also shows a higher accuracy of 76.1\% in \implicitHypernyms{} compared with 61.3\% in \statementOnly{}. This suggests leveraging implicit knowledge is possible in models with non-contextualized representations. However, \esim{} did not achieve perfect accuracy in \softReasoningOnly{}, reaching 79.8\%.  Interestingly, performance on meronymy is similar to hypernymy although no meronyms were provided during training, suggesting the model already has some knowledge of this reasoning skill from pre-training.
%hinting to the robustness of reasoning over different types of implicit knowledge. 

%% file: 04b_analyzing_systematicity.tex
\subsection{Analyzing systematicity} 
\label{subsec:systematicity}

\begin{wraptable}[14]{R}{0.45\columnwidth}
\vspace{-0pt}
\centering
\resizebox{0.45\columnwidth}{!}{
{\small
\setlength\tabcolsep{3pt}    	% default 6pt      
\begin{tabular}{lcc}

Implicit beliefs        & Hypernymy  & \controlledDS{}  \\
\toprule
All correct 			& 99.7      & 95.0     \\
Some incorrect          & 70.0      & 66.3       \\ 
All incorrect          & 13.0      & 7.1       \\ 

\midrule
Overall & 88.8 & 76.9   \\
\midrule
Overall (after intervention)& 98.3 & 94.1   \\
Fraction beliefs corrected & 0.17 & 0.20 \\
\bottomrule
\end{tabular}}
}
\vspace{1mm}
\caption{When the model's implicit beliefs needed for a hypothesis are all correct, scores are high. If we intervene for the small number of incorrect beliefs, adding their rules to the  \context{}, the overall scores increase accordingly.
%\pc{Frac. beliefs incorrect - it's unclear if this is before or after intervention (I know after, but does the reader?) Could possibly move this row to the end of the first block (?). Also perhaps "Intervention overall" $\rightarrow$ "Overall (after intervention)"}}
}
\vspace{0pt}
\label{tab:belief-analysis}
\end{wraptable}

\implicitHypernyms{} achieves 88.8 accuracy, a high, but not perfect, result. A key question is whether the performance in \implicitHypernyms{} can be explained in terms of the implicit knowledge of the LM. To estimate a model's implicit knowledge ``beliefs'', we convert the required implicit knowledge (relevant and distractor hypernyms) to hypotheses without any  \context{} rules (i.e. only \nl{A whale is a mammal} hypothesis), and run the model to check if prediction is correct. We hypothesize that model performance should be high for cases where the implicit beliefs are accurate. 

Table~\ref{tab:belief-analysis} shows how scores depend on whether all the model's implicit beliefs are correct. For the hypernymy examples, the model reaches $99.7$ accuracy when the implicit beliefs are both correct, while it is much lower otherwise. Using the knowledge of which beliefs are incorrect, we can intervene and explicitly add the correct relevant and distractor hypernyms for the small number of incorrect beliefs (17\% in this case). With intervention, the score goes up from $88.8 \rightarrow  98.3$, a significant error reduction.
%JB: deleting for brevity.
%\footnote{As a sanity check, we also find that adding instead in the much larger set of 89\% hypernym rules corresponding to correct beliefs increases the score only to 96.0.}

\comment{
\begin{wrapfigure}{R}{0.4\columnwidth}
\includegraphics[width=0.4\columnwidth]{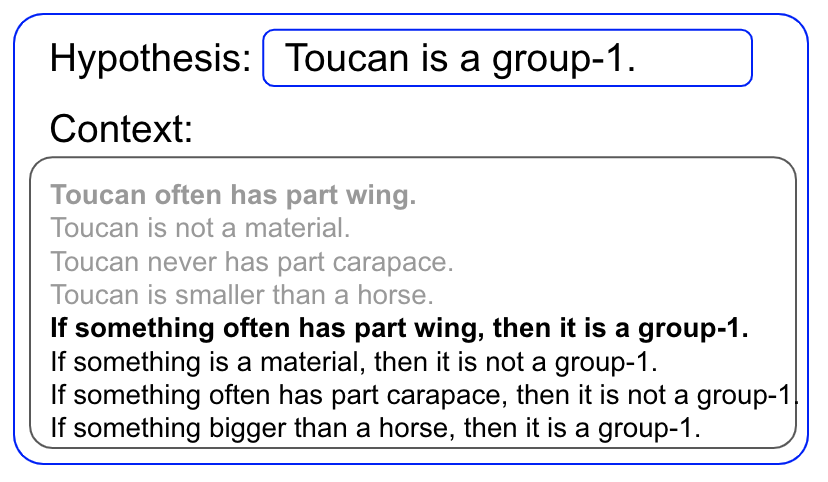}
\caption{Example from \controlledDS{}. \at{i'm really not sure we need this figure. perhaps text is enough and add it to supp?} \jb{if we can keep it let's}}
\label{fig:controlled-example}
\end{wrapfigure}
}

A potential bias in the setup from \S\ref{subsec:taxonomy}, is that the model also has an initial belief about the correctness of the hypothesis itself, regardless of its belief of the hypernym rules. To neutralize this effect we create a controlled dataset \controlledDS{}, where entities have imaginary properties, such as \nl{group-1} in \nl{A Toucan is a group-1}, for which the model does not have any prior knowledge. The rest of the dataset construction is similar to \S\ref{subsec:taxonomy}. Thus, the performance of \statementOnly{} in this setup is 50\% by construction. 
As before there is one relevant implicit rule (\nl{Toucan often has part wing.}), but now there can be up to 5 distractor implicit rules, as well as a mix of rule conditions (hypernyms, meronyms and size comparisons).  In Table~\ref{tab:belief-analysis}, we observe a similar consistency in results when predicting the outcome for the \controlledDS{}-set, where accuracy is $95.0$ when all beliefs are supporting the correct answer, %(this includes incorrect beliefs which happen to imply the correct answer)
$66.3$ when there are conflicting beliefs, and $7.1$ when all beliefs are in support of the wrong answer. 
Intervening here improves overall performance from $76.9 \rightarrow 94.1$.
More details on \controlledDS{} are available in the supp. material. 
To conclude, we observe that model reasoning can be explained by its prior knowledge, and that one can correct it by intervening and correcting false beliefs, suggesting that the model is using implicit knowledge in a systematic fashion.

%with multiple implicit distractor rules. See Figure~\ref{fig:controlled-example} for an example. As before there is one relevant implicit rule ("Toucan often has part wing."), but now there can be up to 5 distractor implicit rules, as well as a mix of rule conditions (hypernyms, meronyms and size comparisons). The explicit rule conclusions are using a controlled language ("is a group-1" or "is not a group-1"), so the \statementOnly{} score is guaranteed to be around 50\%, reducing the amount of confounding factors in this analysis.\footnote{For more more details on the construction and training of the \controlledDS{} dataset see the supplementary material.}

%% file: 04c_counting.tex
%\subsection{Can a Language Model use pre-train knowledge while explicitly counting?}
%\subsection{Approximate Counting over Implicit Facts}
\subsection{Counting over Implicit Facts}
\label{subsec:counting}

\begin{wrapfigure}[11]{R}{0.43\columnwidth}
\vspace{-13pt}
\includegraphics[width=0.43\columnwidth]{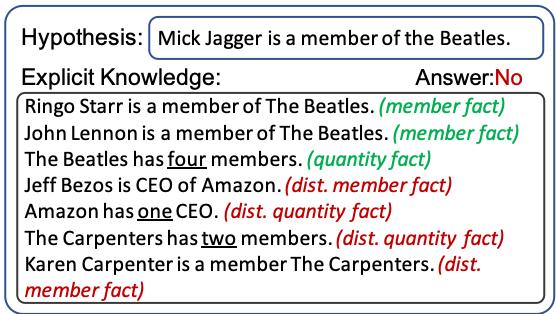}
\caption{Fact counting example outline. The purpose of each fact is in parenthesis. }
\vspace{13pt}
~\label{fig:counting-example}
\end{wrapfigure}

%\pc{Minor: Why counting? The motivation for picking this (vs. other inference types, uncertainty, boolean operators, etc. is unclear.
%perhaps simply switch the first 2 sentences below.}
Our next experiment focuses on whether our models can \emph{simulate} counting over explicit and implicit facts.
While LMs are known to hold taxonomic knowledge \cite{richardson2019does}, there is evidence that skills such as counting are not acquired during pre-training \cite{talmor2019olmpics}. Here, we train a model for simulating counting, and check whether it can count over both text and its prior knowledge. 

\paragraph{Data Construction}
To collect `facts that require counting' (\emph{\memberfact{}s}), we use \textsc{Wikidata} relations such as \nl{member of band}, 
\nl{capital of country}, and \nl{child of person}. \emph{\Quantityfact{}s} provide the total count of facts for a specific entity, e.g., \nl{The Beatles has \textbf{four} members}. Figure~\ref{fig:counting-example} outlines
%shows an outlines of 
a counting example. For each entity (\nl{Beatles}), there are 1-4  \memberfact{}s (\nl{members}).

We create training data by constructing 10,852 \emph{counting sets}, that is, sets that include a single  \quantityfact{} (\nl{The Beatles has four members.}) stating the total number of  \memberfact{}s and $K$  \memberfact{}s,  (\nl{Ringo Starr is a member of The Beatles}), etc.
For each counting set, we create exactly $2K-1$ examples: For all integers $k \in [0, K-1]$, we create an example by adding $k$ member facts and the  \quantityfact{} to the  \context{}, and generating one positive hypothesis and one negative hypothesis. The positive hypothesis is a true member fact not in the  \context{}, and the negative hypothesis is generated by randomly sampling a \texttt{subject} that appears with the relevant \texttt{predicate}, e.g. ((\texttt{Mick Jagger}, \texttt{is member of}, \texttt{The Beatles}), \texttt{false}). When $k=K$, we generate only a false hypothesis. 
%For each number of countable fact added to  \context{}, we generate a positive statement, by sampling one of the countable facts no added to  \context{}, and one negative statement. The negative statement is created by randomly sampling subjects of the same predicate as the countable set at hand, e.g. (\texttt{\nl{Mick Jagger}}, \texttt{\nl{is member of}}, \texttt{\nl{The Beatles}}, false). When all countable facts are added to  \context{}, only a negative statement example is created.

Distractors are added to the  \context{} by adding one  \quantityfact{} and one member fact that share a predicate with the hypothesis, but have a different subject (\nl{The Carpenters}), and one random member fact and \quantityfact{} (\nl{Jeff Bezos is CEO of Amazon}). We find that distractors are important in the sense that if none are added, the model learns to count sentences, and ignores their content, failing to pay attention to the actual subjects of the member facts.

%Distractor countable sets are added to  \context{} by sampling 1-2 countable sets added to other examples with the same predicate. For instance The Carpenters band.  In addition random distractor countable facts are added with different predicates such as (\texttt{\nl{Jeff Bezos}}, \texttt{\nl{is CEO of}}, \texttt{\nl{Amazon}}, true).

Overall 38,700/3,005/3,005 training/development/test examples were created. In 25\% of the examples $k=K$ and thus no implicit knowledge is needed. If the model can count the relevant  \memberfact{}s and verify the count reaches the number specified in the relevant  \quantityfact{}, then it can deduce that the answer is \texttt{false}. In all other examples, solving the task requires combining the  \memberfact{}s that appear in the  \context{} with implicit  \memberfact{}s that the model knows. 

%25\% are deterministic in the sense that all countable facts appear in the  \context{}, thus the answer should be false. Objects of a counted set in training are disjoint from the ones used in evaluation. 

%\at{perhaps mention the distribution of number of counted instance? the vast majority is only 1 counted instance... }

\paragraph{Experiments}
We evaluate our approach using \roberta{} and \esim{} (the performance of \zeroshot{} is low since it cannot count without training).
\begin{itemize}[leftmargin=*,topsep=0pt,itemsep=2pt,parsep=0pt]

\item[] \countingstatementonly: 
The model is given only the hypothesis (similar to \S\ref{subsec:taxonomy}).
%without the  \context{}. This tests whether the model knows the answer without the need for any inference.

\item[] \countingexperiment: Examples are sampled from the same distribution as the training set. In 25\% of the examples the model can use explicit counting to predict \texttt{false} ($k=K$). In the rest, it must combine prior knowledge of  \memberfact{}s with member facts that are in the  \context{}.    
\end{itemize}

%Here, we introduce a set of tests and baselines designed to test for weather the LM can utilize pre-trained knowledge.

%\countingstatementonly: Here we test the models ability to correctly predict the statement by removing all relevant rules in the examples.

%\countedfacts: Similar to training set examples. In 25\% of the examples the model can use explicit counting to predict False. In the rest of the example it must combine it's pretrained knowledge with the existing facts in  \context{} to predict the answer.    

\paragraph{Results}

\begin{wraptable}[11]{R}{0.43\columnwidth}
\vspace{-10pt}
\begin{center}
\resizebox{0.43\columnwidth}{!}{
\begin{tabular}{ll|c}
Experimental setup                 & Subset & \roberta    \\
\toprule
\countingstatementonly      &  $(1,K-1)$    &  64.1      \\
%                            &  $K$        &  -        \\
\countingexperiment               &  $(1,K-1)$    &  73       \\
                            &  $K$        &  99.7     \\
\bottomrule
\end{tabular}}
\end{center}
\caption{Test set performance for counting. We show performance on two subsets of the test set: $(1,K-1)$: where the number of  \memberfact{}s is in $1, \dots, K-1$, and $K$ where the number of member facts is exactly $K$.   }
\vspace{10pt}
\label{tab:counting}
\end{wraptable}

%We evaluate the experiments on three models, one fine-tuned on the counting training data, a \zeroshot{} model \at{that was described earlier?} and \esim{}.

% \zeroshot  results: 55.6, 53, 96.2 

%Because examples require knowledge of facts that are rarely reported, we use \roberta{} that has a relatively higher amount of implicit knowledge. 
Table~\ref{tab:counting} shows the accuracy for all models. We distinguish between the case where the total number of  \memberfact{}s has been reached ($k=K$), where the answer is \texttt{false}, and the rest of the examples.
\roberta{} achieves a near perfect result of 99.7\% when $k=K$, illustrating it can be trained to count the relevant member facts.\footnote{We
also conduct an experiment where we provide $K$ member facts, but some of them are false (e.g., \nl{Mick Jagger is a member of the Beatles}) and find that the model simply counts member facts, ignoring whether they are factually correct or not, still predicting \texttt{False} in all such cases.}

When the number of  \memberfact{}s is in $1, \dots, K-1$,
accuracy improves by $9$ points given the \context{} ($64.1 \rightarrow 73$), implying that the model learned to combine implicit knowledge of member facts with the \context{}.
Both \zeroshot{} (not trained on the counting dataset) and \esim{}, trained on the generated data, predicted \texttt{false} to almost all examples, indicating they are not well suited for this task.
To further validate that the model is indeed counting, we dropped the relevant quantity fact from all test examples.  The accuracy drops from $73 \rightarrow 64.4$ (Table 3, counting $(1, K-1)$), which is similar to hypothesis-only $64.1$, suggesting that the model is using the quantity fact to know how many member facts should be counted.

Performance shows that the model uses the \context{} to improve prediction. However, because the required implicit knowledge involves long-tail facts that might be missing from the LM, learning to use this knowledge can be difficult. We perform a more fine-grained analysis to decouple prior knowledge from the counting task at hand.

\begin{wrapfigure}[15]{R}{0.53\columnwidth}
\vspace{-17pt}
\includegraphics[width=0.53\columnwidth]{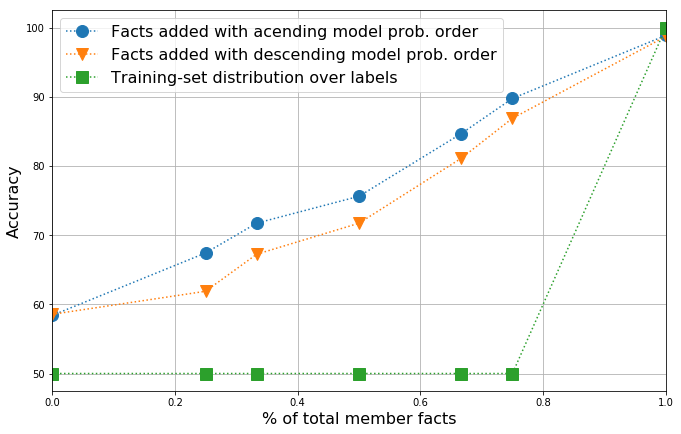}
\caption{An analysis of \roberta{} accuracy (y-axis) vs $c=\frac{k}{K}$ (x-axis).}~\label{fig:counting-analysis}
\vspace{17pt}
\end{wrapfigure}

\paragraph{Analysis}
Our analysis setup is the following. We take all \memberfact{}s, and obtain a \emph{fact probability}, which is the output of our model when given the  \memberfact{} as a hypothesis without any \context{}. Then, for each counting set we create $2K-1$ examples as before, but for each $k \in [1\dots K-1]$ we take $k$ facts not \emph{randomly}, but according to one of two orders: (a) ascending fact probability, and (b) descending fact probability. We hypothesize that if  \memberfact{}s are sorted in ascending order, performance should be higher, because the model is given explicit evidence for facts it is less confident about. 

%To further investigate weather \roberta{} can combine implicit knowledge and explicit counting, we in add  \memberfact{}s to the  \context{} incrementally, from zero to total number of  \memberfact{}s. 
%However, as opposed to adding the  \memberfact{}s in random order as in the previous experiment, here we order appending the  \memberfact{}s depending on the probably in which \roberta{} predicts these facts are true. We evaluate this probability for each  \memberfact{} separately.

%We consider to orders of addition: ascending probabilities and descending probabilities. 
Figure~\ref{fig:counting-analysis} shows the results. The x-axis represents $c=\frac{k}{K}$, that is, the fraction of  \memberfact{}s given in the  \context{}. When $c=0$ or $c=1$, ascending and descending orders are equivalent, since either no facts or all facts are given. Green squares show the distribution over labels for each value of $c$, illustrating that guessing based on label leads to 50\% accuracy, except when $c=1$, in which all examples are \texttt{false}.
%Results are displayed in Figure~\ref{fig:counting-analysis}. The X axis represents the portion of  \memberfact{}s appearing in the  \context{} from 0 (no relevant  \memberfact{}s) to 1 (all  \memberfact{}s appear). The green squares show the training set gold answer majority distribution for each portion, exactly 50\% for each value except 1.0 to which the answer is 100\% false.
Blue circles display accuracy when  \memberfact{}s are added in ascending order, and orange triangles in descending order. 

When $c=0$, model accuracy is slightly lower than 60\%, showing that \roberta{} has non-trivial knowledge of the relevant background facts. When $c=1$ performance is perfect since our models count the correct facts and reach $K$. 
For both ascending and descending orders, the accuracy of  \roberta{} monotonically increases with the number of  \memberfact{}s, although distribution over labels is constant. This suggests that explicit  \memberfact{}s help the pre-trained LM leverage its prior knowledge. When $c=0.75$ both models hover around an impressive 90\% accuracy.
We confirm our hypothesis that adding facts in ascending order substantially improves performance: when $c=0.25$ performance improves by 5.5 points, and when $c=0.75$ it improves by 2.7.
%In addition, we notice that adding  \memberfact{}s in ascending probability order achieves consistently higher accuracy (77.9\% on average) than when added in descending order (73.4\% on average) in portions between 0 and 1. 
This shows that adding  \memberfact{}s which the model does not ``know'' improves performance more than a fact it is already
confident about. 
%that the model is combining implicit knowledge with explicit  \memberfact{}s present in the  \context{}. Thus adding a  \memberfact{} the model does not "know", helps it more than one it is confident about. Implying that in some cases it is able to implicitly take into account the  \memberfact{} it is more confident about.
%\at{ we can add the Beatles Ringo vs Paul example as intuition here, but i'm not doing this yet, because we are starting to have space issues... } \jb{yeah add it.} \at{no more room, let's talk about what we would like to add ... figure?}

%% file: 04d_combining_skills.tex
\subsection{Generalizing to New Skill Combinations}
In \S\ref{subsec:taxonomy} and \S\ref{subsec:counting}, we isolated a single inference type in each experiment. However, our goal is to have a model that can seamlessly combine these skills. To test this, we train models on various subsets from \ruletaker{}, \twentyquestions{} (described in \S\ref{sec:motivation}), \counting{} (\S\ref{subsec:counting}) and \hypernyms{} (\S\ref{subsec:taxonomy}). We then test whether these models can handle examples that require multiple skills simultaneously.
Training is done by randomly mixing the datasets and performing standard multi-task training.

%To test this, we combine training data from different skills, to fine-tune on one model. Examples from each training data-set were mixed randomly, and the model was fine-tuned in multi-task.

\begin{wrapfigure}[11]{r}{0.42\columnwidth}
\vspace{-2pt}
\includegraphics[width=0.42\columnwidth]{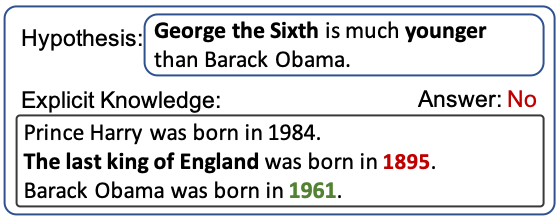}
\caption{\manualset{} example, requiring age/year + hypernymy skills.}
\vspace{2pt}
~\label{fig:multi-skill-set-example}
\end{wrapfigure}

Ultimately, we want users to correct models on-the-fly by mixing natural language statements that demand different reasoning skills.
To emulate this process, we created \manualset{}, composed of 185 hand-crafted hypotheses and \context{},\footnote{All examples and best model prediction are available in the supplementary material.} labeled with the set of skills needed for each example. Figure~\ref{fig:multi-skill-set-example} shows one example, requiring age/year comparison + hypernymy skills.
We specifically chose hypotheses \roberta{} answers incorrectly in \statementOnly{}.

\paragraph{Results} Table \ref{tab:combining_skills} shows results on \manualset{}.
The rows show the subset of datasets the model was trained on and the columns show the accuracy for examples that require a particular skill.
A model trained on all 4 datasets obtains an accuracy of $85.4$.
All models perform poorly given only the hypothesis, with an average accuracy of $40.2$. (Less than random, due to the adverserial manner in which we chose examples for \manualset{}). 
Models trained without \counting{} show low accuracy on the counting skill subset, suggesting that this skill was not acquired at pre-training time.
Interestingly, models combining \hypernyms{} or \counting{} with \ruletaker{} and \twentyquestions{}, display higher accuracy than models trained only on one of these datasets. This suggests that models are able to successfully combine multiple reasoning skills, some newly acquired \cite{Talmor2019MultiQAAE}.
Models show high accuracy on skills that they are not explicitly trained for, but have been shown to be acquired during pre-training, such as size comparison \cite{talmor2019olmpics}, and meronyms \cite{richardson2019probing}. 

The right side of Table \ref{tab:combining_skills}, presents the accuracy on examples requiring at least two skills. Strikingly, the model is able to accurately compose multiple skills, reaching a high accuracy of $88$ on examples combining implicit hypernyms and counting.  Size, age, and year comparison examples are not available in our training sets, nevertheless, models achieve an impressive score of $81.8$ and $73.9$ on examples combining implicit hypernyms with size and age/year, respectively.  

\begin{table}[h]
\begin{center}
\resizebox{1\columnwidth}{!}{
\begin{tabular}{l|cc|cccc|ccc}
test questions, skills $\rightarrow$ & overall & hypothesis & hypernyms  & counting & sizes & age/year & hypernyms & hypernyms & hypernyms  \\
training data $\downarrow$ &         & only     &  &  &        &   & + counting & + sizes & + age/year \\
\toprule
\hypernyms{}                & 62.7          & 50.8          & 64.7          & 34.0 & 69.2 & 46.7 & 28.0 & 72.7 & 43.5 \\
\counting{}                 & 66.5          & \textbf{54.6} & 65.5          & 72.3 & 46.2 & 63.3 & 60.0 & 45.5 & 69.6 \\
\zeroshot{}                 & 69.2          & 34.6          & 76.5          & 55.3 & 73.1 & 60.0 & 68.0 & 68.2 & 60.9 \\
\hypernyms{}+\zeroshot{}    & 73.5          & 31.9          & 74.8          & 44.7 & 61.5 & 76.7 & 48.0 & 63.6 & \textbf{73.9} \\
\counting{}+\zeroshot{}     & 70.8          & 34.6          & 68.1          & \textbf{83.0} & 65.4 & 56.7 & \textbf{88.0} & 59.1 & 52.2 \\
\textsc{All combined}       & \textbf{85.4} & 34.6          & \textbf{84.0} & \textbf{83.0} & \textbf{84.6} & \textbf{80.0} & 84.0 & \textbf{81.8} & \textbf{73.9} \\
\bottomrule
\end{tabular}}
\end{center}
\caption{Accuracy for \manualset{}. Rows show training set composition for \roberta{}. Overall accuracy, hypothesis-only accuracy, single-skill and multiple-skill breakdown are displayed in columns.   }
\label{tab:combining_skills}
\end{table}

%% file: 05_related_work.tex
\section{Related Work}

We have demonstrated that LMs can be taught to {\it systematically} combine both pre-trained knowledge and explicit natural language statements. This predictability is important: If the model is behaving rationally, we can teach it.
This distinguishes our work from other research in recognizing textual entailment (RTE),
e.g., \cite{Dagan2013RecognizingTE,Bowman2015ALA, Parikh2016ADA, Tay2018CompareCA},
where models learn to score a hypothesis but in an opaque and somewhat unpredictable way
\cite{Gururangan2018AnnotationAI,Nie2019AdversarialNA}.
It similarly distinguishes our work from multi-hop QA, e.g., \cite{yang2018hotpotqa,talmor2018web,welbl2017constructing, Asai2020LearningTR, Xiao2019DynamicallyFG}, where again model behavior is often opaque \cite{Min2019CompositionalQD}.

% Similarly, our work is qualitatively distinct from work on multihop QA,
% e.g., \cite{yang2018hotpotqa}. Again, for these problems, rules of
% inference (i.e., which multihop chains are valid) are opaque and
% need to be inferred from examples (e.g., that ``causes'' is transitive).
% In our case, we 

There have been numerous earlier demonstrations that neural systems can learn systematic behavior,
including for semantic parsing \cite{Li2019CompositionalGF},
symbolic integration \cite{Lample2019DeepLF}, mathematics \cite{Saxton2019AnalysingMR}, knowledge prediction \cite{petroni2020context,petroni2019language},
and rule-based reasoning \cite{clark2020transformers}. However, these
are all largely self-contained tasks. We extend this to show how
implicit knowledge can be directly harnessed in a systematic inference process.

Finally, although teaching a machine via general statements has long been a goal of AI \cite{Mccarthy1959ProgramsWC},
current neural methods typically require large numbers of examples to convey knowledge
\cite{Hu2016HarnessingDN}. Our work shows how pre-trained networks can instead be 
taught on-the-fly using a few {\it general} statements, in a ``one-shot'' manner that exploits pre-trained knowledge, without requiring re-training of the model.

%% file: 06_discussion.tex
\section{Discussion}

%\jb{Another attempt - shorter} \at{fine-tuning a bit}
In this work, we show that pre-trained LMs can be trained to consistently combine implicit knowledge encoded in their parameters with explicit rules and facts. Models are able to perform various types of reasoning in this setup including multi-hop reasoning, counting, number comparisons, and taxonomic knowledge. 
Moreover, we show that one can inject the ability to perform various types of inferences one at a time independently, and obtain generalization to cases that require combining these skills in a single example. Our work opens the door to models that learn through interaction with users. Users can teach the model facts and rules about the world through natural language statements, and the model will utilize this new information immediately, combining it with the knowledge encoded internally. Such an approach allows users to ``teach the model'' and correct its current and future errors without the need for data collection and re-training, in a one-shot manner.

%% file: 07_statement_of_broader_impact.tex
\section*{Broader Impact}
% https://medium.com/@operations_18894/a-guide-to-writing-the-neurips-impact-statement-4293b723f832

Our work, if successful, paves a possible path towards models that learn in a one-shot manner by interacting with users. Users of Question Answering and other systems utilizing reasoning over natural language may benefit by constantly improving models that do not require re-training in an interactive manner.
However, users teaching false rules and facts may lead to the spread of ungrounded and possibly ``fake'' information. Thus, the provided rules and facts must be constantly monitored and curated. 
Finally, users relying on the reasoning of such systems for mission critical tasks, such as medical advice, might be at risk from possible errors. At current level of accuracy of state-of-the-art models, this type of usage is not advised.

%Authors should discuss both positive and negative outcomes, if any. For instance, authors should discuss a) 
%who may benefit from this research, b) who may be put at disadvantage from this research, c) what are the consequences of failure of the system, and d) whether the task/method leverages
%biases in the data. If authors believe this is not applicable to them, authors can simply state this.

%Use unnumbered first level headings for this section, which should go at the end of the paper. {\bf Note that this section does not count towards the eight pages of content that are allowed.}

%% file: 08_acks.tex
\begin{ack}
We thank our colleagues at The Allen Institute of AI and Tel-Aviv University, especially Kyle Richardson, Nicholas Lourie, Ashish Sabharwal, Elad Segal, Mor Geva and Tomer Wolfson.
This research was partially supported by The Israel Science Foundation grant 942/16, The Blavatnik Computer Science Research Fund and The Yandex Initiative for Machine Learning, and the European Union’s Seventh Framework Programme (FP7) under grant agreement no. 802774-ERC-iEXTRACT and no. 802800-DELPHI.
\end{ack}

%% file: 10_supplementary.tex
\clearpage
\section{Supplementary Material}
\label{sec:supp}

\subsection{Method}
Pseudo-language statements are generated using manually constructed templates for each \texttt{predicate}. The predicates used include the 
\textsc{ConceptNet} relations \texttt{(IsA, Antonym, DistinctFrom, PartOf, CapableOf, Desires, NotDesires)}. In addition, \texttt{(IsHypernymOf, IsMeronymOf)} relations were produced by combining \textsc{wordnet} and \textsc{ConceptNet}, for the Hypernymy and Meronymy datasets.

For the \counting{} task, we use the following predicates for the \memberfact{}s: \texttt{(super bowl winner, super bowl loser, band member, capital, director, release year, founder, headquarter, child, spouse,CEO)}.
For \quantityfact{}s, predicates include \texttt{(has 1,..,has 5)}, effectively supporting up to a count of five \memberfact{}s.

\subsection{Implicit Knowledge of Taxonomic Relations}

\paragraph{Experiments}
We evaluate our model in two additional setups:
\begin{itemize}[leftmargin=*,topsep=0pt,itemsep=2pt,parsep=0pt]
\item[] \statementOnlyNoContext:
We compare the effect of removing the \context{} entirely verses the \statementOnly{} setup, in which only the relevant rules are removed. 
\item[] \statementOnlyLangSelectivity:
To insure the model is using the hypothesis subject, i.e. \nl{Whale} in \nl{A whale has a belly button}, we replace the subject of the statement with random words that carry no meaning such as \nl{foo, blah, ya, qux, aranglopa, foltopia, cakophon, baz ,garply}. This experiment controls for weather the model can answer correctly solely based on the context.
\end{itemize}

\paragraph{Results} Table~\ref{tab:hypernym-meronym-cont} compares the two new experiments introduced.  \statementOnlyNoContext{} shows slightly higher results on average, suggesting that the distractors in the \context{} partially mislead the model compared to a case where no \context{} is used.

\begin{table}[h]
\begin{center}
\resizebox{0.9\columnwidth}{!}{
\begin{tabular}{l|c|c|c|c}
Model  $\rightarrow$ & \multicolumn{3}{c|}{\roberta{}}    & \esim{}  \\
\toprule
Train-set $\rightarrow$  & Hypernymy & Hypernymy  & \zeroshot{} & Hypernymy  \\ 
\midrule
Test-set $\rightarrow$ & Hypernymy & Meronymy   & Hypernymy & Hypernymy   \\
\midrule
\statementOnly                          &   65.2    &  70.8     &  65.4     &  61.3      \\
\statementOnlyNoContext                 &   66.7    &  71.1     &  68.0     &  59.0     \\
\statementOnlyLangSelectivity           &   55.6    &  55.5     &  57.7     &  54.9     \\
\bottomrule
\end{tabular}
}
\end{center}
\caption{Test set results for reasoning over hypernymy and meronymy relations. The models learn to reason with implicit rules, significantly improving on the hypothesis-only baseline, some in zero-shot.}
\label{tab:hypernym-meronym-cont}
\end{table}

In the \statementOnlyLangSelectivity{} experiment, the model achieves a slightly higher than random accuracy of ~55\%, implying that the \context
{} does give away the answer to a small number of examples but in the  rest the full hypothesis is needed to arrive at the correct answer.

\subsection{Analyzing systematicity} 
\paragraph{Details on \controlledDS{} dataset} 
In the "real world rules" experiments presented in Section~\ref{subsec:taxonomy}, the model may also have a prior belief about the {\it answer} from pre-training, adding a potentially confounding element, e.g., is the model right because of reasoning or prior knowledge?

\begin{wrapfigure}[13]{R}{0.45\columnwidth}
\includegraphics[width=0.45\columnwidth]{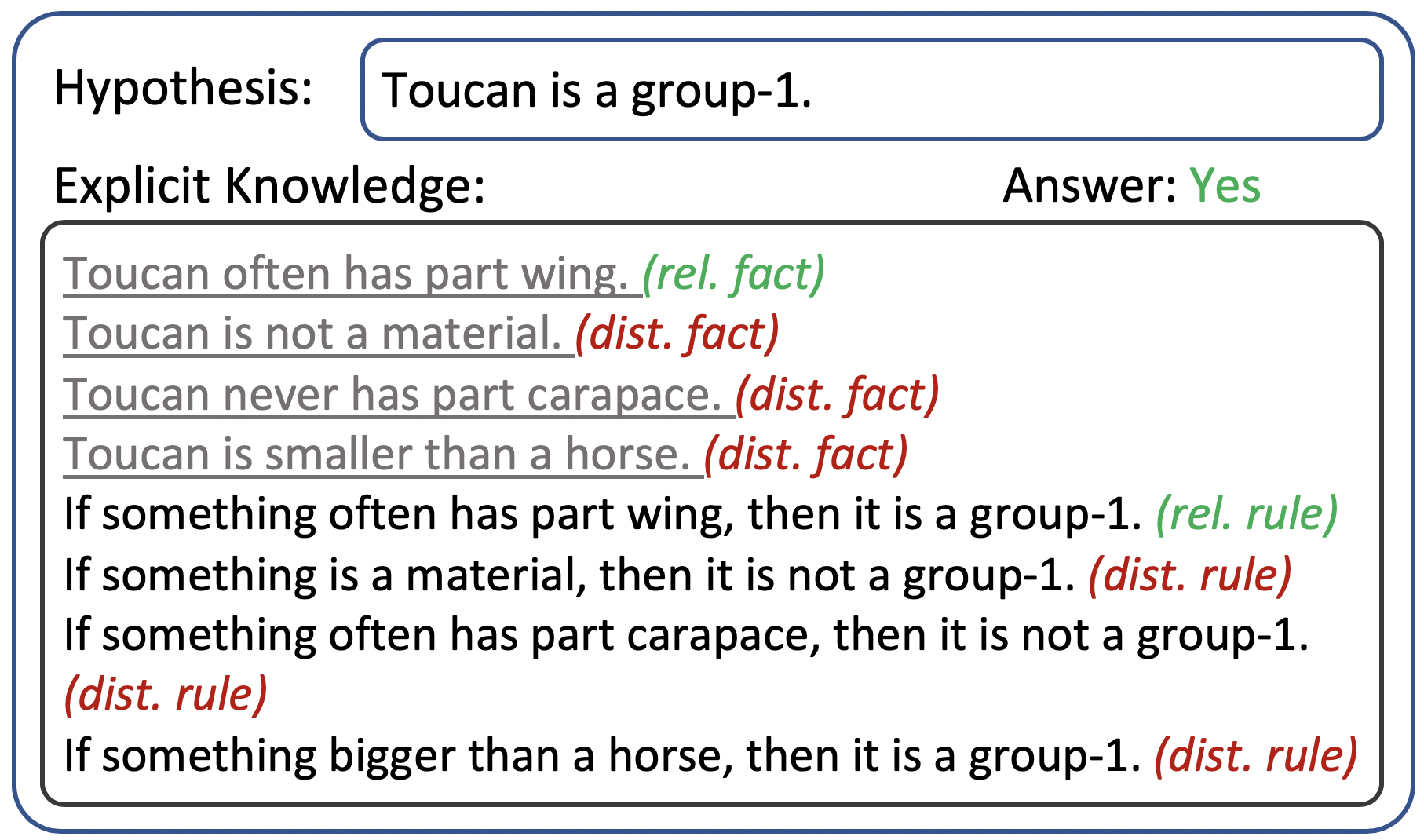}
\caption{Example from the \controlledDS{} dataset. Underlined facts are removed in \implicitHypernyms{}. }
\label{fig:controlled-example}
\end{wrapfigure}

To factor this out, we perform another experiment using the \controlledDS{} dataset (Figure~\ref{fig:controlled-example}) with neutral rule conclusions of the form \nl{X is a group-N}. We generate rule sets where, for the entity in the hypothesis, only one condition is true (relevant fact) while 3-5 others are false (distractor facts). We divide the rules randomly into positive and negative rules. We mix three different relation types (hypernym, meronym, size comparison), and distractors are sampled to be somewhat adversarial (inverse relation, other relation, etc). For the \implicitHypernyms{} setting, we exclude all the facts from the context.

We train a \roberta{} model as in Section~\ref{subsec:taxonomy}, on a mix of settings (\softReasoningOnly{}, \implicitHypernyms{}, and\statementOnly{} for the knowledge facts). The development and test sets use a disjoint hypernym tree of entities from the training set. As expected, this model scores near-random (52.3\%) on the \statementOnly{} variant. It scores near-perfect on \softReasoningOnly{} (99.0\%), while scoring 79.2\% on the \implicitHypernyms{} variant.

\subsection{Generalizing to New Skill Combinations}

To further explore the systematicity of reasoning over multiple skills, we create \nl{templates} for which arguments are replaced with multiple values and the answer is updated accordingly. For each template we automatically generate dozens of examples, and compare the accuracy of \statementOnly{} with the effect of adding the  \context{}. We evaluate the examples on the \textsc{all combined} model. Results, shown in Table \ref{tab:templates}, display an average increase of 56\% in accuracy when adding the  \context{}, achieving a high accuracy of 93.5\% on average. This suggests that the model consistently combines multiple implicit skills, such as hypernymy, age comparison, year comparison, as well as explicit multi-hop reasoning. 

\begin{table}[h]
\begin{center}
\resizebox{1\columnwidth}{!}{
\begin{tabular}{l|l|cc}
{}                      &  {}                 & hypothesis  & with      \\
hypothesis +  \context{} template    &  template arguments         & only        &  \context{}    \\
\midrule
H: \textcolor{blue}{[NAME]} can live up to \textcolor{ForestGreen}{[LIFE\_SPAN]} years.                   & \textcolor{blue}{[NAME]}: John, Elizabeth, Chen, Richard, Don, Moses & 50.0 & 90.0 \\ 
C: A Human can live to up an older age than a \textcolor{BrickRed}{[ANIMAL\_TYPE]}.        & \textcolor{ForestGreen}{[LIFE\_SPAN]}: 5,10,70,80 & &\\
A whale can live up to 200 years.                                   & \textcolor{BrickRed}{ANIMAL\_TYPE}, \textcolor{Purple}{LIFE\_SPAN}: (Horse:30), (Panda:20),& & \\
A \textcolor{BrickRed}{[ANIMAL\_TYPE]} can live up to \textcolor{Purple}{[ANIMAL\_LIFE\_SPAN]}{} years.       & (Cow:22), (Monkey:30), (Tiger:15)& & \\
\midrule
H: \textcolor{red}{[ENTITY]} contains \textcolor{blue}{[PARTICLE]}. & \textcolor{blue}{[PARTICLE]}: Protons, Atoms,  Neutrons, Quarks & 25.0 & 96.9 \\ 
C: A physical entity is made of matter. Matter contains \textcolor{blue}{[PARTICLE]}.        & \textcolor{red}{[ENTITY]}: A Frog, A Car, A Mountain,  & &\\
Energy does not contain \textcolor{blue}{[PARTICLE]}. & A River, Light, Radiation, Electricity, Magnetism  & & \\
Abstract entities do not contain \textcolor{blue}{[PARTICLE]}. &  & & \\
\bottomrule
\end{tabular}}
\end{center}
\caption{An analysis of \textsc{All combined} performance over a set of examples generated from templates. The template hypothesis and  \context{} are displayed on the left column. Template arguments, enclosed in brackets, with their respective values in the middle column. Hypothesis-only, and hypothesis with  \context{} accuracy, are shown on the right. }
\label{tab:templates}
\end{table}